\documentclass[10pt,twocolumn,letterpaper]{article}

\usepackage{iccv}
\usepackage{times}
\usepackage{epsfig}
\usepackage{graphicx}
\usepackage{amsmath}
\usepackage{amssymb}

\usepackage[breaklinks=true,bookmarks=false]{hyperref}

\iccvfinalcopy 

\def\iccvPaperID{****} 

\ificcvfinal\fi

\begin{document}

\title{Searching for Accurate Binary Neural Architectures}

\author{Mingzhu Shen \quad Kai Han \quad Chunjing Xu \quad Yunhe Wang\\
Huawei Noah's Ark Lab\\
{\tt\small \{shenmingzhu,kai.han,xuchunjing,yunhe.wang\}@huawei.com}
}

\maketitle
\ificcvfinal\thispagestyle{empty}\fi

	\begin{abstract}
	Binary neural networks have attracted tremendous attention due to the efficiency for deploying them on mobile devices. Since the weak expression ability of binary weights and features, their accuracy is usually much lower than that of full-precision (\ie 32-bit) models. Here we present a new frame work for automatically searching for compact but accurate binary neural networks. In practice, number of channels in each layer will be encoded into the search space and optimized using the evolutionary algorithm. Experiments conducted on benchmark datasets and neural architectures demonstrate that our searched binary networks can achieve the performance of full-precision models with acceptable increments on model sizes and calculations.
\end{abstract}

\section{Introduction}

Convolutional neural networks (CNNs) have been widely used in various computer vision tasks, such as image classification~\cite{resnet}, object detection~\cite{Ren2015Faster} and visual segmentation~\cite{long2015fully}. These neural networks are often of heavy design with massive parameters and computational costs, which cannot be directly deployed on portable devices without model compressing techniques, \eg pruning~\cite{deepcompression}, knowledge distillation~\cite{distill}, compact model design~\cite{mobilenet,wang2018learning}, and quantization~\cite{xnor,dorefa}.

\begin{figure}[t]
	\centering
	\includegraphics[width=0.9\linewidth]{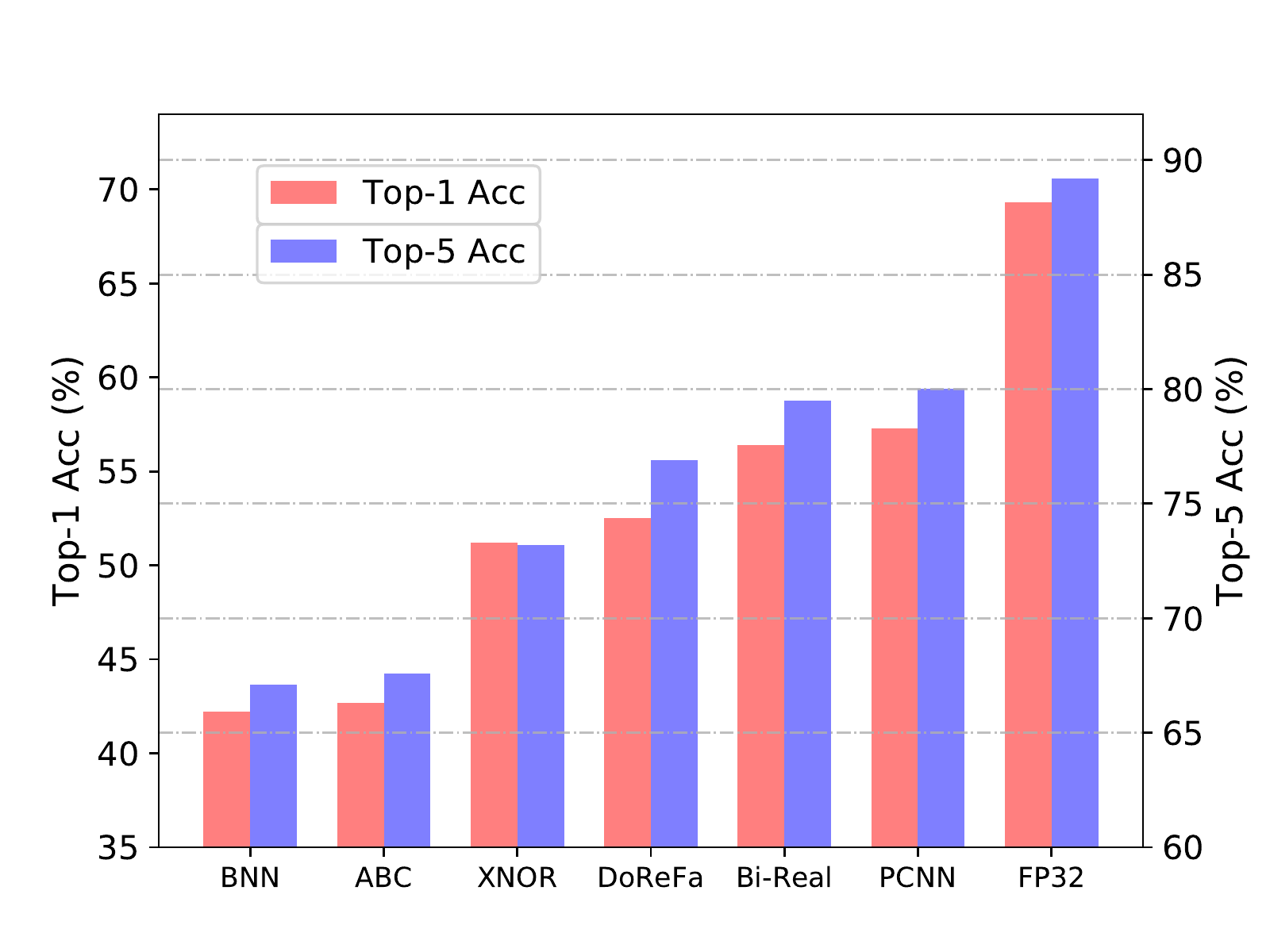}
	\vspace{-0.5em}
	\caption{Performance of state-of-the-art methods for binarizing ResNet-18 on the ImageNet dataset.}
	\label{bnn}
	\vspace{-1.5em}
\end{figure}

Wherein, 1-bit quantization has been recently received a great attention, which represents the weights and activations in the network using only two values, \eg $-1$ and $+1$. Thus, binarized networks could be efficiently applied in a series of real-world applications (\eg camera and mobile phone). Nevertheless, the performance of binary neural networks (BNNs) are still far worse than that of their original models. Figure~\ref{bnn} summarizes the performance of state-of-the-art binarization methods~\cite{bnn,abc,xnor,dorefa,bi-real,PCNN} on the ImageNet benchmark~\cite{imagenet}, including XNOR-Net~\cite{xnor}, Bi-Real Net~\cite{bi-real}, PCNN~\cite{PCNN}, \etc. Although they have made tremendous efforts for enhancing the performance of BNNs, the highest top-1 accuracy obtained by PCNN~\cite{PCNN} is about $12.0\%$ lower than that of the baseline ResNet-18~\cite{resnet}.

The severe accuracy drop mentioned in Figure~\ref{bnn} greatly limits the practicality of BNNs, considering that there are a number of computer vision taks with very high precision requirements such as face recognition~\cite{deepid} and person re-identification~\cite{han2019attribute}. The main reason could be derived from the fact that discrimination of binary features cannot match that of the full-precision features with the same dimensionality. Therefore, it is necessary to find a trade-off approach for establishing compact binary networks with acceptable model sizes by increasing the number of channels in each convolutional layer. Motivated by the recent neural architecture search (NAS~\cite{nas,evonas,wang2017towards}) hotspot, we present to appropriately modify channel numbers of binarized networks and search a new architecture with different channel numbers but high precision. In practice, expansion ratios of all layers in the desired binary network will be encoded to form the search space, and the evolutionary algorithm will be utilized for effectively find the lower bound of BNNs for achieving the same performance as that of their full-precision versions.

We conduct experiments on the CIFAR and ImageNet datasets using VGGNet~\cite{vggnet} and ResNet~\cite{resnet} architectures. Results on these benchmarks show that the proposed approach is able to find excellent binary neural architectures for obtaining high precision with as few computation costs as possible.

\section{Approach}

\paragraph{Binarization Method.}
Following the widely-used DoReFa-Net~\cite{dorefa}, in the binary layer, the floating-point weights $\mathbf w$ is approximated by binary weights $\mathbf w_{b}$ and a floating-point scalar, while the floating-point activations $\mathbf x$ are represented by binary values $\mathbf x_{b}$. The feed-forward in DoReFa-Net is defined as:
\begin{equation}
\begin{aligned}
\mathbf w_{b} &= \text{sign}(\mathbf w) \times E(|\mathbf w|), \\
\mathbf x_{b} &= \text{round}(\text{clip}(\mathbf x,0,1)),
\end{aligned}
\end{equation}
where $E(|\cdot|)$ calculates the mean of absolute value. In the back-propagation process, we adapt the ``Straight-Through Estimator'' method~\cite{ste} to estimate the corresponding gradients. During the quantization process, we restrain the weights and activations of all convolution layers and fully-connected layers to only 1-bit except the first and last layer, following the existing works~\cite{dorefa,bi-real}.

The extremely binary quantization brings enormous computation acceleration and memory reduction. However, most of the state-of-the-art binary networks cannot match the accuracy of the full-precision counterpart models. Recently, the uniform width expansion proposed by WRPN~\cite{wrpn} expands all the layers with only one hyper-parameter for multi-bit quantization networks to pursue this goal. 

Although widened binary networks can obtain acceptable performance, such a uniform expansion strategy will obviously increase the required memory and computational complexities, \eg the binary network after expanding $4\times$ is $16\times$ larger than the original one. In fact, there is often strong redundancy in deep neural architectures, we do not need to expand all layers for achieving the desired performance. Thus, we propose to define a binary neural architecture search problem and utilize evolutionary algorithm to search the optimal architectures.

\paragraph{Search Space.}
For the search space, we only focus on the search for network width, \ie the number of the channels of each layer. For a given network architecture which has $n$ layers, we define $\mathbf a \in \mathbf R^{n}$ to encode the expansion ratio hyper-parameter of each layer. Our goal is to search $\mathbf a$ for higher accuracy with less FLOPs. All the other hyper-parameters and network settings like stride, kernel size, layer order, remain the same as the original full-precision models.

In the uniform width expansion experiments as shown in Table~\ref{resnet18}, we observe that by only expanding channels by $4$ times, binary neural networks can obtain comparable performance to that of their full-precision model on the ImageNet classification task. Thus we assume that $4$ is the empirical upper bound of expansion ratio to achieve full-precision accuracy. We set $4$ as the largest expansion ratio, and use some smaller ratio to expand or even reduce channels. In practice, we have $6$ expansion ratio candidates in $\mathbf a$ which is defined as follows:
\begin{equation}
\mathbf a = [a_1,...,a_n],\quad\forall\;\; a_i \in\{0.25,0.5,1,2,3,4\}.
\label{ratiocode}
\end{equation}	

\paragraph{Search Algorithm.}	
As discussed above, we expect to search an optimal architecture with the expansion ratio set $\mathbf a^{\ast}$ for making the accuracy of the binarized neural networks similar to that of its full-precision models with as few parameters and floating-number opeartions (FLOPS) as possible. Therefore, the overall optimization can be described as:
\begin{equation}
\begin{aligned}	
\max_{\mathbf{a}}\quad& f(\mathbf w^{\ast}(\mathbf a),\mathbf a),\\
s.t. \quad & \mathbf w^{\ast}=arg\min_{\mathbf w}\mathcal{L}_{train}(\mathbf w, \mathbf a),
\end{aligned}
\end{equation}
where $f(\cdot)$ is the \emph{fitness} function in evolutionary algorithm and $\mathcal{L}_{train}$ is loss on train set, $\mathbf w^{\ast}(\mathbf a)$ is the corresponding trained weight with expansion ratio set $\mathbf a$. We first find an optimal $\mathbf a^{\ast}$ through evolutionary algorithm on a train subset. Then we train the corresponding binary network on full train set to obtain the final model.

Specifically, in every generation during evolution, we maintain a population of $K$ individuals, \ie $\{\mathbf a_1,...,\mathbf a_K\}$, each of which denotes a bianry neural architecture according to a certain expansion ratio code satisfying Eq.~\ref{ratiocode}. These individuals will be continuously updated with pre-designed operations (\eg corssover and mutation) to have greater fitness. Here we have two objects: high performance on the specific task, \eg classification accuracy, and low computation costs, \eg FLOPs. Thus, the fitness $f(\mathbf a_k)$ of an individual $\mathbf a_k$ is defined as:

\begin{equation}
\begin{aligned}
f(\mathbf a_k) &= \max(\text{Acc} - \lambda \times \text{FLOPs}, 0)
\end{aligned}
\label{fitness}
\end{equation}
where $\text{Acc}$ and $\text{FLOPs}$ are the Top-1 validation accuracy and FLOPs of the corresponding widened networks of the individual $\mathbf a_k$, $\lambda$ is the trade-off parameter.

Compared with full-precision layers, the FLOPs of binary layers are divided by $64$ as suggested in Bi-Real Net~\cite{bi-real}. In the calculation of fitness in Eq.~\ref{fitness}, we divide the FLOPs of the candidate models by the FLOPs of original binary network to get the same order of magnitude of accuracy. After defining the search space and fitness function, the evolutionary algorithm can effectively select excellent individuals with higher fitness during the evolution process until convergence.

\section{Experiments}
In this section, we conduct experiments to explore the empirical width lower bound of each layer in binary neural networks on several benchmark datasets, \ie CIFAR-10~\cite{cifar}, and ImageNet~\cite{imagenet}. We use two widely used network structures as baselines, VGG-small~\cite{hwgq} and ResNet-18~\cite{resnet}.

\subsection{Experimental Settings}
For the evolution search process, we search for 50 generations with 32 individuals in each generation. We train each candidate model for 10 epochs on the trainset and obtain the accuracy on validation set as the accuracy used in Eq.~\ref{fitness}. For the trade-off parameter $\lambda$, we set it to $4$ to keep the value of accuracy and FLOPs comparable. 

\paragraph{CIFAR-10}
In CIFAR-10 dataset, it takes about 12 hours on 8 V100 GPUs. Then we train 200 epochs for full CIFAR-10 training. The learning rate starts as 0.1 and multiply by 0.1 in the epochs of 60, 120 and 180. We simply follow the same hyper-parameter setup as that in~\cite{hwgq}.

\paragraph{ImageNet}
As the ImageNet ILSVRC2012 dataset is very large, we do not use the whole train dataset in evolution process. We randomly sample a subset of 50,000 images from the original full trainset which belongs to 1000 classes with 50 images for each class in the evolution process and it takes about 180 hours on 8 V100 GPUs. Then we train 150 epochs to check if searched models reaches full-precision accuracy. The learning rate starts from 0.1 and decays by 0.1 in the epochs of 50, 100 and 135. We simply follow the same hyper-parameter setup as that in~\cite{resnet}.	

\paragraph{Initialization}
When evaluating each candidate, we train 10 epochs on a small subset in ImageNet dataset, the accuracy of candidate models is especially low and makes it difficult to distinguish the better models from the worse ones. Therefore, we train the model uniformly widened by $4\times$ on the subset with 150 epochs and use it to initialize all the candidate models which we simply intercept first corresponding channels values.

\begin{table}[h]
	\centering
	\small
	\caption{Comparison of widened binary networks of VGG-small architecture on CIFAR-10.}
	\renewcommand\arraystretch{1.0}
	\begin{tabular}{|c||c|c|c|c|}
		\hline
		Models&FLOPs&Speedup&Memory&Top-1(\%)\\
		\hline\hline
		Full-Precision&608M&-&149M&\textbf{93.48}\\
		\hline
		Uniform-1$\times$&13.2M&46.1$\times$&7.3M&90.24\\
		Uniform-2$\times$&45.3M&13.4$\times$&23.7M&91.65\\
		Uniform-3$\times$&96.2M&6.3$\times$&49.3M& 91.87\\
		Uniform-4$\times$&166M&3.7$\times$&84.1M&92.56\\
		\hline
		VGG-Auto-A&11.3M&53.6$\times$&5.1M&92.17\\
		VGG-Auto-B&59.3M&10.3$\times$&23.4M&\textbf{93.06}\\	
		\hline
	\end{tabular}
	\label{vgg-small}
\end{table}
\subsection{Results and Analysis}

\paragraph{VGG-small on CIFAR-10}
VGG-small~\cite{hwgq} is a variant network of the original VGG-Net~\cite{vggnet} designed for CIFAR-10. We compare the searched models, \ie Automatic-A, B, with uniformly widened models in Table~\ref{vgg-small}. The standard binarized VGG-Small decreases accuracy only by about $3\%$. As we uniformly increase the width, the accuracy increases subsequently. However with 4$\times$ widened, the accuracy of binarized network still does not achieve that of full-precision network. Our Automatic-B model achieves higher accuracy than the Uniform-4$\times$ with about 1/4 FLOPs and memory. It has the smallest accuracy gap with the full-precision model. Although our Automatic-A model even has less channels than the original Uniform-1$\times$ model, it achieves higher accuracy with about 2\% improvement. This phenomenon confirms our original intention in designing the search space, that some layers need to be expanded and some layers need to be narrowed. 

\begin{table}[h]
	\centering
	\small
	\caption{Comparison of widened binary networks and other binarization methods of ResNet-18 architecture on ImageNet dataset.}
	\renewcommand\arraystretch{1.0}
	\begin{tabular}{|c||c|c|c|c|}
		\hline
		Models&FLOPs&Speedup&Top-1(\%)&Top-5(\%)\\
		\hline\hline
		Full-Precision&1820M&-&\textbf{69.6}&89.2\\
		\hline
		PCNN&169M&10.8$\times$&57.3&80.0\\
		ABC$\{5/3\}$&520M&3.5$\times$&62.5&84.2\\
		ABC$\{5/3\}$&785M&2.3$\times$&65.0&85.9\\
		\hline
		Uniform-1$\times$&149M&12.2$\times$&52.77&76.85\\
		Uniform-2$\times$&352M&5.2$\times$&64.0&85.45\\
		Uniform-3$\times$&607M&3.0$\times$&68.51& 88.25\\
		Uniform-4$\times$&915M&2.0$\times$&70.35&89.27\\
		\hline
		Res18-Auto-A&495M&3.7$\times$&68.64&88.46\\
		Res18-Auto-B&660M&2.8$\times$&\textbf{69.65}&89.08\\	
		\hline
	\end{tabular}
	\label{resnet18}
\end{table}

\paragraph{ResNet-18 on ImageNet}
We also conduct experiments on the large-scale ImageNet dataset. In the uniform expansion experiments, as the width increases, the top-1 accuracy can gradually approach that of the original full-precision model. From the results in Table~\ref{resnet18}, our Automatic-B binarized model can obtain the the same performance with the full-precision model with less than 1/3 computational cost. With similar FLOPs, Automatic-B outperform Uniform-3$\times$ by 1.1\% in terms of Top-1 accuracy and 0.8\% Top-5 accuracy. Our evolutionary search finds a more accurate widened models with as less FLOPs as possible.

We also compare our models with some state-of-the-art binarization methods in Table~\ref{resnet18}. PCNN~\cite{PCNN} does not quantize the downsample layer and adds additional shortcut connections which could inevitably increase end-to-end inference time. In the comparison of ABC-Net with multiple bases, which $5/3$ means 5 binary bases for weight and 3 bases for activations, Our Uniform and Automatic models consistently performs better than ABC-Net by a large margin. 

\begin{figure}[htb]
	\centering
	\small
	\begin{tabular}{cc}
		\includegraphics[width=1\linewidth]{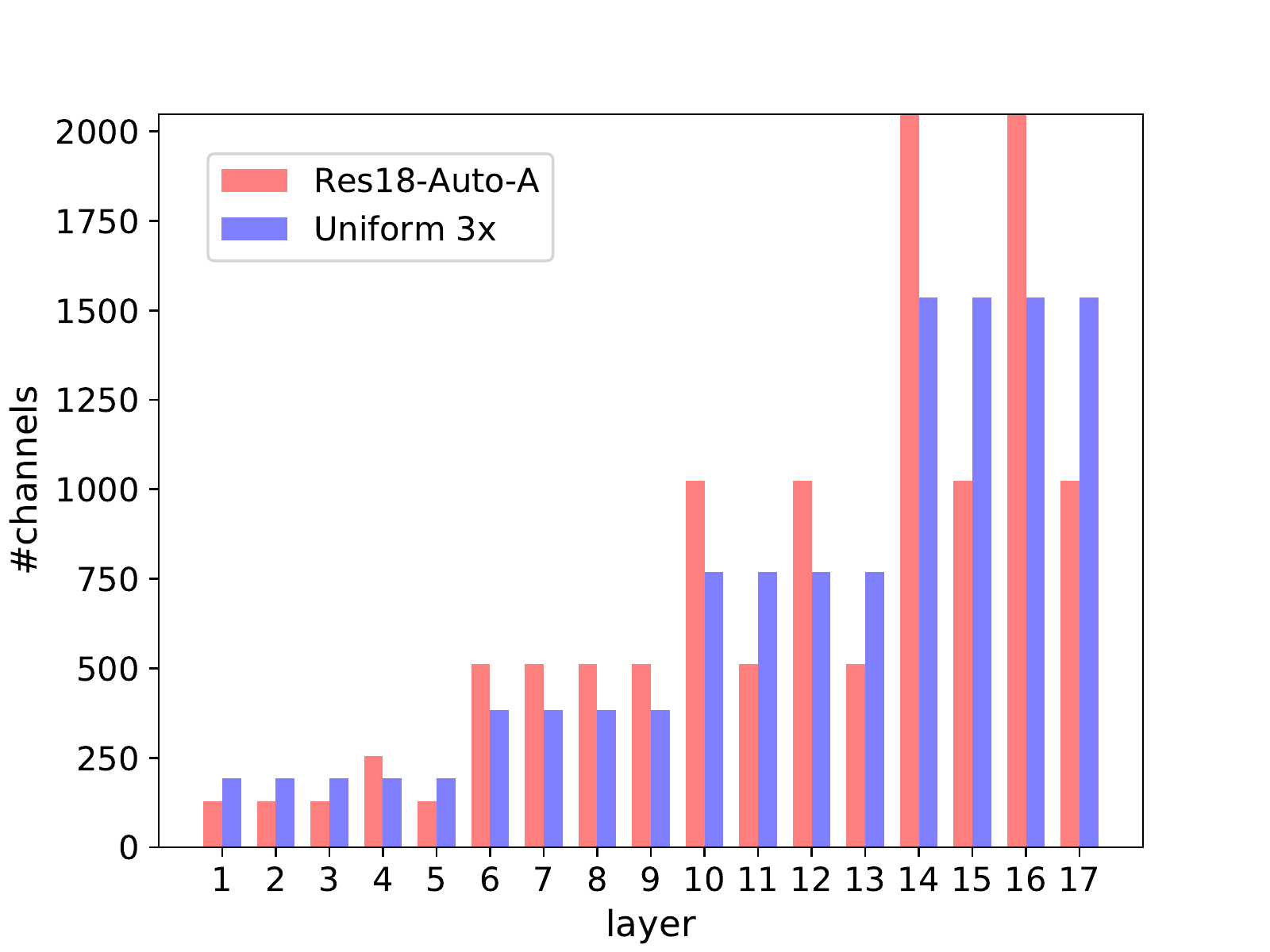}
	\end{tabular}
	\caption{Number of channels in each layer of widened ResNet-18.}
	\label{resnet18-vis}
\end{figure}
\paragraph{Searched Architecture}
To further analyze the searched network architecture, we show the number of output channels in each layer of two binary networks with similar accuracy, \ie Res18-Auto-A and Uniform-3$\times$ in Table~\ref{resnet18}. From Fig.~\ref{resnet18-vis}, we observe that compared with Uniform-3$\times$, the searched architecture Res18-Auto-A has fewer output channels in the 1st, 2nd and last stages. In addition, Res18-Auto-A needs more channels for the middle feature maps inside each block. These observations could inspire us to design blocks or architectures for more efficient convolutional neural networks.

\section{Conclusion}
To establish binary neural networks with higher precision and lower computational costs, this paper studies the binary neural architecture search problem. Based on the empirical study on uniform width expansion, we define a novel search space and utilize evolutionary algorithm to adjust the number of channels in each convolutional layer after binarizing. Experiments on benchmark datasets and neural architectures show that the proposed method can produce binary networks with acceptable parameters increment and the same performance as that of the full-precision original network.

{\small
	\nocite{han2018autoencoder}
	\bibliographystyle{ieee}
	\bibliography{egbib}
}

\end{document}